\documentclass[letterpaper]{article} % DO NOT CHANGE THIS
\usepackage{aaai25}  % DO NOT CHANGE THIS

\usepackage{times}  % DO NOT CHANGE THIS
\usepackage{helvet}  % DO NOT CHANGE THIS
\usepackage{courier}  % DO NOT CHANGE THIS
\usepackage[hyphens]{url}  % DO NOT CHANGE THIS
\usepackage{graphicx} % DO NOT CHANGE THIS
\usepackage{natbib}  % DO NOT CHANGE THIS AND DO NOT ADD ANY OPTIONS TO IT
\usepackage{caption} % DO NOT CHANGE THIS AND DO NOT ADD ANY OPTIONS TO IT
\usepackage{colortbl}

\usepackage{algorithm}
\usepackage{algorithmic}
\usepackage{amsmath}

\usepackage{newfloat}
\usepackage{listings}
\DeclareCaptionStyle{ruled}{labelfont=normalfont,labelsep=colon,strut=off} % DO NOT CHANGE THIS
\floatstyle{ruled}
\newfloat{listing}{tb}{lst}{}
\floatname{listing}{Listing}
\title{
Towards Scientific Discovery with Generative AI: Progress, Opportunities, and Challenges
}
\author{
 Chandan K Reddy, \ Parshin Shojaee
}
\affiliations{
   Virginia Tech\\
    reddy@cs.vt.edu, \ parshinshojaee@vt.edu
}

\usepackage{bibentry}
\begin{document}
\maketitle

\begin{abstract}
Scientific discovery is a complex cognitive process that has driven human knowledge and technological progress for centuries. While artificial intelligence (AI) has made significant advances in automating aspects of scientific reasoning, simulation, and experimentation, we still lack integrated AI systems capable of performing autonomous long-term scientific research and discovery. This paper examines the current state of AI for scientific discovery, highlighting recent progress in large language models and other AI techniques applied to scientific tasks. We then outline key challenges and promising research directions toward developing more comprehensive AI systems for scientific discovery, including the need for science-focused AI agents, improved benchmarks and evaluation metrics, multimodal scientific representations, and unified frameworks combining reasoning, theorem proving, and data-driven modeling. Addressing these challenges could lead to transformative AI tools to accelerate progress across disciplines towards scientific discovery.
\end{abstract}

\section{Introduction}
% \ps{a comprehensive figure is needed for both recent advances as well as challenges and research potentials}

Scientific discovery - the process of formulating and validating new concepts, laws, and theories to explain natural phenomena - is one of humanity's most intellectually demanding and impactful pursuits. For decades, AI researchers have sought to automate aspects of scientific reasoning and discovery. Early work focused on symbolic AI approaches to replicate the formation of scientific hypotheses and laws in symbolic forms \citep{segler2018planning,maccoll1897symbolic}. More recently, deep learning and large language models (LLMs) have shown promise in tasks like literature analysis and brainstorming \cite{ji2024scimon,lu2024ai,si2024can}, experiment design \cite{boiko2023autonomous,arlt2024meta}, hypothesis generation \cite{wang2024hypothesis,ji2024scimon}, and equation discovery \cite{shojaee2024llm,llmsim-bilevel}.

\begin{figure}[t]
\centering
\includegraphics[width=\columnwidth]{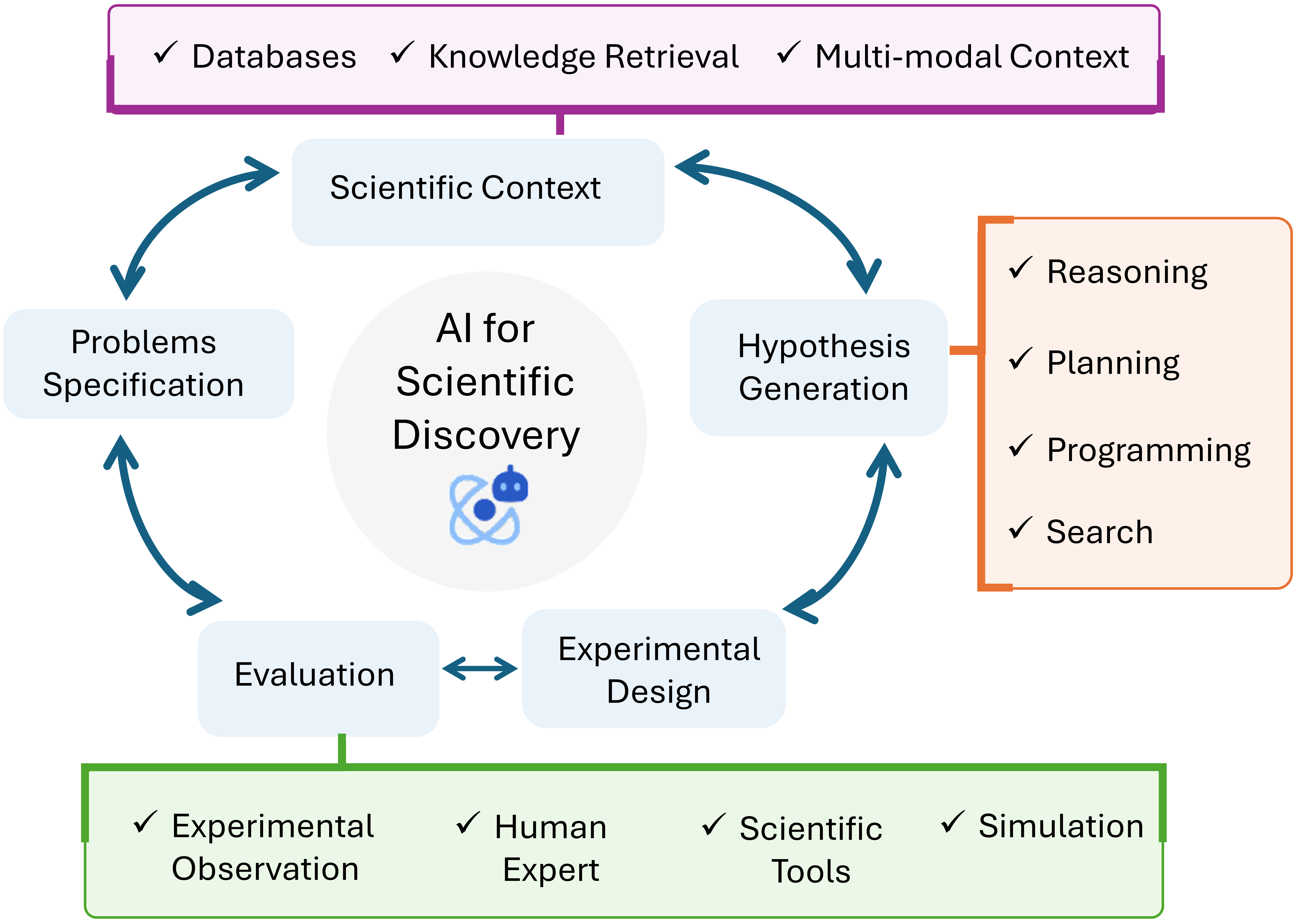}
\vspace{-0.5em}
\caption{
% \small
\textbf{Overview of the AI-driven scientific discovery framework.} The cycle illustrates the 
% integration of large generative models in the 
iterative process of scientific inquiry.
% , encompassing problem specification, context retrieval, hypothesis generation, experimental design, and evaluation.
The framework begins with user-defined problem specifications, retrieves relevant scientific context from literature and databases, and utilizes generative AI systems to produce new hypotheses and experimental designs.
These AI-generated concepts are then evaluated and refined through experimental observation, expert input, and scientific tools, driving further iterations of the discovery cycle.
\vspace{-0.5em}
% (in natural language or representation space). 
% Overview Framework of the AI for Scientific Discovery Framework. This framework is an iterative hypothesis refinement process with these steps: problems specification, sceitnfic context, hypothesis generation, experiment design, and evaluation. In this framework, first, scientific problem specification is received from user, then relevant scientific context is retrieved from literature and databases, these context along with problems specification is fed to generative AI systems, then generative AI systems offer new hypothesis and experiments designs (in natural language by LLMs or in representation space), these are hypotheses and designs are then 
}
\label{fig:scidisc}
\vspace{-1.0em}
\end{figure}

Despite this progress, we still lack AI systems capable of integrating the diverse cognitive processes involved in sustained scientific research and discovery. Most work has focused on narrow aspects of scientific reasoning in isolation. 
Developing more comprehensive AI discovery systems capable of supporting the full cycle of scientific inquiry —from context retrieval and hypothesis generation to experiment design and evaluation (Figure \ref{fig:scidisc}) —could dramatically accelerate progress across scientific disciplines.
This paper examines the current state and future potential of generative AI for scientific discovery. We highlight recent advances, particularly in scientific understanding and discovery frameworks, while identifying critical gaps.
We then outline key research challenges and directions towards more unified AI systems for discovery, including: ($i$)~Creating improved benchmarks and evaluation frameworks for scientific discovery; ($ii$)~Developing science-focused AI agents that leverage scientific knowledge and reasoning capabilities; ($iii$)~Advancing multimodal scientific representations beyond text; and ($iv$)~Unifying automated reasoning, theorem proving, and data-driven modeling. 
By tackling these challenges, the AI and Science community can work towards systems that serve as collaborative partners to human scientists, accelerating the pace of discovery in science.
% like climate science \cite{}, biomedicine \cite{}, material science \cite{}, and physics \cite{}.

\section{Recent Advances in AI for Scientific Tasks}
The past decade has witnessed remarkable progress in applying AI to various scientific tasks. This section highlights some of the most significant recent advances, demonstrating AI's growing capabilities in supporting and accelerating scientific discovery across multiple disciplines.

\subsection{Literature Analysis and Brainstorming}
The exponential growth of scientific publications has made it increasingly challenging for researchers to stay abreast of developments in their fields. Large language models (LLMs) pre-trained on vast scientific corpora have emerged as powerful tools to address this challenge, enhancing literature analysis and interaction. Researchers have developed specialized LLMs for various scientific domains. Models like PubMedBERT \cite{gu2021domain} and BioBERT \cite{lee2020biobert} focus on biomedical literature, while SciBERT \cite{beltagy2019scibert} covers a broader range of scientific disciplines. More recent models such as BioGPT \cite{luo2022biogpt} and SciGLM \cite{zhang2024sciglm} have further pushed the boundaries of scientific language modeling, incorporating advanced architectures and training techniques. 
These models, trained on sources like PubMed and arXiv, excel at literature information retrieval, summarization, and question-answering. They enable efficient navigation of scientific knowledge by quickly finding relevant papers, distilling key findings, and synthesizing information to answer complex queries.

Beyond analysis, recent works demonstrate LLMs' potential in generating novel scientific insights. For instance, \text{SciMON}~\cite{ji2024scimon} uses LLMs to generate new scientific ideas by analyzing patterns in the existing literature. These advancements show AI's capacity to not only aid in literature review but also contribute to identifying promising and novel research directions, potentially accelerating scientific discovery.

\subsection{Theorem Proving}
Automated theorem proving has recently gained attention in AI for science research due to its fundamental role in scientific reasoning. Recent years have seen remarkable progress in this field, particularly through the integration of LLMs with formal reasoning systems.
The GPT-f framework \cite{polu2020generative} pioneered this approach by training transformer-based language models on proof tactics, enabling navigation through complex mathematical proofs with the help of learned priors. Building on this, researchers have integrated proving techniques with LLMs and developed enhancements such as data augmentation \cite{han2021proof}, retrieval augmentation \cite{yang2024leandojo}, and novel proof search methods \cite{lample2022hypertree,wang2023dt}.
One of the key enhancements is the autoformalization approach, exemplified by the Draft-Sketch-Prove method \cite{jiang2023draft}. This method uses LLMs to first draft informal proofs, translate them into formal sketches, and then complete proofs with additional proof assistant tools \cite{bohme2010sledgehammer}, mimicking the human process of moving from intuitive understanding to rigorous proof.
As these systems become more adept at formalizing and proving complex statements, they could be applied to derive scientific theories, potentially accelerating the scientific process and leading to enhancements in fields where theoretical understanding lags behind empirical methods.

\subsection{Experimental Design}
Experimental design is a critical component of the scientific process, often requiring extensive domain knowledge and creative thinking. The automation of this process through generative models has the potential to accelerate scientific discovery across various fields.
By leveraging LLM agents, researchers are recently developing systems that can design, plan, optimize, and even execute scientific experiments with minimal human intervention. These tools are particularly valuable in fields where experimental setup is costly, allowing researchers to explore a wider range of possibilities before physical implementation.
For example, in physics, LLM-driven systems have demonstrated effectiveness in designing complex quantum experiments \cite{arlt2024meta} and optimizing parameters in high-energy physics simulations  \cite{cai2024transforming,baldi2014searching}.
Chemistry has also recently seen advancements in automated experimentation, with LLM agent systems capable of designing and optimizing chemical reactions \cite{m2024augmenting}. 
Moreover, in biology and medicine, LLM-driven experimental design has shown promise in optimizing
% CRISPR 
gene-editing protocols \cite{huang2024crispr}, and designing more effective clinical trials \cite{singhal2023large}.
These AI-driven approaches to experimental design allow researchers to tackle more complex problems and explore hypotheses that might otherwise be impractical due to time or resource constraints.

\subsection{Data-driven Discovery}
Data-driven discovery has become a cornerstone of modern scientific research, leveraging the ever-growing volumes of experimental, observational, and synthetic data to uncover new patterns, relationships, and laws.
% that might elude traditional hypothesis-driven approaches.
This paradigm shift has been particularly transformative in fields where complex systems and high-dimensional data are prevalent. 
% Examples of these domains include: 

In drug discovery, data-driven approaches have significantly accelerated the identification of potential therapeutic compounds. For instance, recent works employed generative \cite{mak2023artificial,callaway2024major} and multi-modal representation learning \cite{gao2024drugclip} models to discover a novel antibiotic, effective against a wide range of bacteria, by searching and screening millions of molecules in the representation space \cite{gao2024drugclip}. These enhancements demonstrate the power of AI in exploring vast chemical spaces that would be infeasible to search manually or in the huge and infinite combinatorial space of molecules.

Equation discovery, commonly known as symbolic regression, is a data-driven task for uncovering mathematical expressions from data.  
% This field has evolved rapidly, incorporating diverse approaches from physics-inspired algorithms to advanced AI techniques.
Early neural methods like AI Feynman \cite{aifeynman1} demonstrated the ability to rediscover fundamental physics laws from data alone, while later work incorporated physical constraints and structures for more interpretable models \cite{cranmer2020discovering}. 
% Advanced evolutionary search models such as PySR \cite{cranmer2023interpretable} further expanded the search capabilities in the vast space of mathematical functions.
The advent of language modeling and representation learning brought new possibilities. Transformer-based language models, adapted for symbolic regression, treat equation discovery as a numeric-to-symbolic generation task \cite{biggio2021neural,kamienny2022end}. These approaches have been enhanced with search techniques during decoding \cite{landajuela2022unified,shojaee2024transformer}, although challenges remain in effectively encoding and tokenizing numeric data \cite{golkar2023xval}.
Recent works like the SNIP model \cite{meidani2024snip} have also explored multi-modal representation learning between symbolic expressions and numeric data, moving the equation discovery search to a lower-dimensional and smoother representation space for more effective and efficient search. 
Recently, LLM-SR \cite{shojaee2024llm} also demonstrated the potential of using LLMs as scientist agents in the evolutionary search for equation discovery.
These advancements highlight the evolving landscape of equation discovery, with significant potential for further improvements in integrating numeric data with AI models and leveraging the mathematical reasoning capabilities of advanced LLMs.

In materials discovery, data-driven approaches have led to the prediction and subsequent synthesis of novel materials with desired properties \cite{pyzer2022accelerating,merchant2023scaling,miret2024llms}. 
Large generative models have shown remarkable success in generating novel structures. For instance, \citet{merchant2023scaling} introduced Graph Networks for Materials Exploration (GNoME), leading to the discovery of new stable materials. This approach represents an order-of-magnitude increase in known stable crystals, showcasing the potential of AI in expanding our materials knowledge base.
LLMs have also been recently used to extract information from scientific literature in material science, generate novel material compositions, and guide experimental design \cite{miret2024llms}.
For example, the AtomAgents \cite{ghafarollahi2024atomagents} demonstrates how LLMs can be integrated into the material discovery pipeline, significantly improving the process in alloy design.
By combining the pattern-recognition and representation learning capabilities with the reasoning and generalization abilities of advanced AI models, we are moving towards systems that can not only analyze existing data but also propose novel hypotheses for data-driven discoveries across scientific disciplines.

\begin{figure*}[t]
\centering
\includegraphics[width=0.75\textwidth]{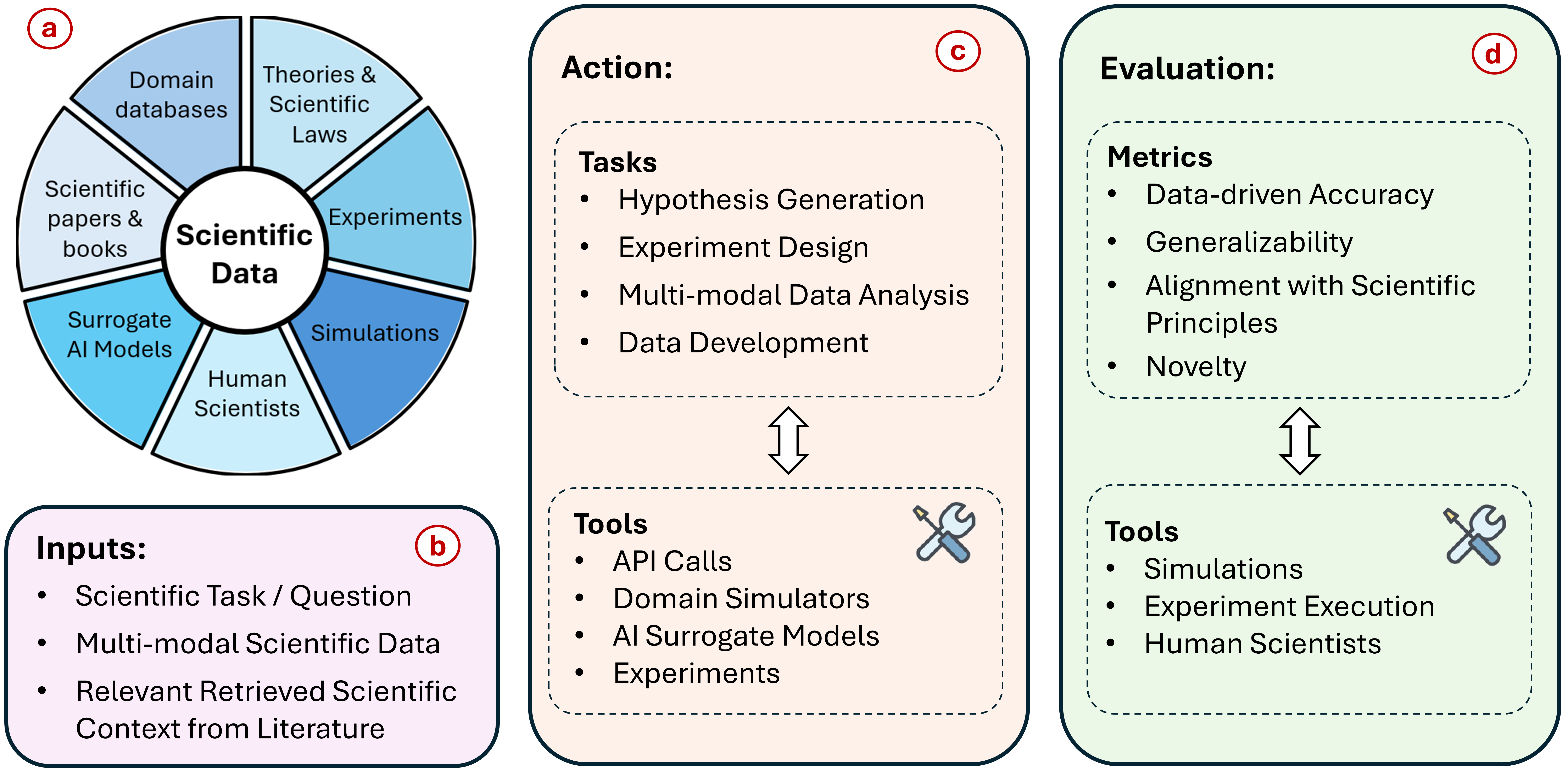}
% \vspace{-0.5em}
\caption{
% \small
A comprehensive framework for \textbf{science-focused AI agents}. The diagram illustrates \textbf{\textcircled{a}}~the multi-modal nature of scientific data, \textbf{\textcircled{b}}~the inputs for scientific tasks, \textbf{\textcircled{c}}~the key actions performed by AI agents in scientific discovery, and \textbf{\textcircled{d}}~the evaluation metrics for assessing scientific outcomes. This framework highlights the integration of diverse data sources, AI-driven tools, and human experts in advancing scientific research and discovery processes.
}
\label{fig:sciagent}
% \vspace{-0.5em}
\end{figure*}

\section{Key Challenges and Research Opportunities}

\subsection{Benchmarks for Scientific Discovery}
First and foremost, evaluating AI systems for open-ended scientific discovery poses unique challenges compared to typical machine learning benchmarks. This challenge is particularly acute for large language models (LLMs) and other foundation models capable of storing and potentially ``memorizing" vast amounts of scientific knowledge \cite{brown2020language,bommasani2021opportunities} in their parameters. 
Many existing benchmarks in the field of scientific discovery only focus on rediscovering known scientific laws or solving textbook-style problems. For instance, the AI Feynman dataset consists of 120 physics equations to be rediscovered from data \cite{aifeynman1,aifeynman2}, while datasets like SciBench \cite{wang2023scibench}, ScienceQA \cite{scienceqa}, and MATH \cite{hendrycks2021measuring} primarily evaluate scientific question answering and mathematical problem-solving abilities.

However, these benchmarks may not capture the entire complexity of scientific discovery processes. More critically, they may be vulnerable to reciting or memorization by large language models, potentially leading to overestimation of true discovery capabilities \cite{carlini2021extracting,shojaee2024llm}. As \cite{wu2023reasoning} points out, LLMs can often solve scientific problems by pattern matching against memorized knowledge rather than through genuine reasoning or discovery. This concern is further emphasized by studies showing that LLMs can reproduce significant portions of their training data \cite{carlini2022quantifying}.
There is a pressing need for richer benchmarks and evaluation frameworks in this research area to better understand the gap between baselines and recent methods and to identify areas for improvement. Key directions include:

\begin{itemize}
\item \textit{Developing benchmark datasets focused on novel scientific discovery rather than recovery}:
One promising approach is to create configurable simulated scientific domains where the underlying laws and principles can be systematically varied. This would allow testing discovery capabilities on new scenarios, mitigating the risk of models simply reciting memorized information observed in their training data. For example, \cite{m2024augmenting} used a simulated chemistry environment to evaluate AI-driven discovery of novel chemical reactions. 
Similarly, \cite{shojaee2024llm} designed simulated settings for different scientific domains such as material science, physics, and biology to evaluate AI-driven scientific equation discovery. 
A key challenge in this line of research is balancing the use of LLMs' prior scientific knowledge while avoiding mere recitation or memorization. This balance is crucial for advancing AI's role in scientific discovery.

\item \textit{Creating evaluation metrics for multiple facets of scientific discovery}: 
To comprehensively assess scientific discovery capabilities, we need a multi-faceted evaluation framework. The key metrics include:
($i$)~\textit{Novelty}: 
Measures to quantify how different a discovered hypothesis or law is from existing knowledge. This could involve comparing against a corpus of known scientific literature \cite{ji2024scimon};  
($ii$)~\textit{Generalizability}: Assessing how well discovered laws or models predict out-of-distribution unobserved data. To do so, evaluation benchmarks should be developed that test discovered laws on scenarios significantly different from the training data distribution, highlighting how scientific theories should be generalizable to new contexts; 
($iii$)~\textit{Alignment with Scientific Principles}: Evaluating whether discovered hypotheses are consistent with fundamental laws of physics or other well-established scientific knowledge. This could involve developing formal verification methods for scientific consistency \cite{cornelio2023combining,cranmer2020lagrangian}, as well as assessing the discovered laws' compatibility with existing scientific theories \cite{liu2024kan}.

\item \textit{Involving domain experts in benchmark design and evaluation}: 
The involvement of domain experts is crucial for developing meaningful benchmarks and evaluating AI-driven scientific discoveries. 
Experts can contribute in various aspects of the discovery process such as assessing the plausibility, novelty, and potential impact of AI-generated hypotheses; 
evaluating the interpretability and alignment of AI-discovered laws or models with human-understandable scientific principles; and providing feedback during the AI-driven discovery process for human-AI collaborative discovery.
By integrating domain expert involvement throughout the benchmark development, discovery, and evaluation process, we can ensure that advancements in AI-driven scientific discovery are both technically sound and aligned with the needs and standards of the scientific community.
\end{itemize}

\subsection{Science-Focused Agents}
Current work on scientific AI often treats models as passive tools rather than active agents pursuing discovery. There is a growing need to develop science-focused AI agents (Figure \ref{fig:sciagent}) that can leverage broad scientific knowledge, engage in reasoning, and autonomously verify their reasoning and hypotheses. Recently, LLMs have shown impressive capabilities in knowledge retrieval and reasoning \cite{huang2023towards}, making them promising candidates for developing such agents. These agents can integrate vast amounts of scientific knowledge embedded in LLMs, generate educated hypotheses, design experiments, verify their designs, and interpret the results. Also, their ability to interface with external tools and experimental data sources with the programming execution gate allows for real-world experimentation and validation. 
Recent work has demonstrated the potential of LLM-based agents in scientific domains. For example, \cite{m2024augmenting} introduced ChemCrow, an LLM-augmented system for chemistry research. ChemCrow integrates GPT-4 with domain-specific tools for tasks such as reaction prediction, retrosynthesis planning, and safety assessment. This integration allows the system to reason about chemical processes and validate the hypotheses using specialized chemical tools.
Similarly, \cite{ghafarollahi2024atomagents} developed AtomAgents, a multi-agent system for alloy design and discovery. 
SciAgents \cite{ghafarollahi2024sciagents} also uses multiple AI agents, each specializing in different aspects of materials science, to collaboratively design new bio-materials. The system incorporates physics-aware constraints and can interface with simulation tools to validate its predictions.
% , demonstrating how AI agents can navigate complex scientific domains and generate testable and educated hypotheses.
However, developing effective science-focused agents also presents several challenges:
% The motivation for developing LLM-based scientific agents stems from their potential to accelerate scientific discovery by automating research sub-tasks.

\begin{itemize}
\item \textit{Domain-specific tool integration}: 
Effective scientific agents require integration with specialized scientific tools and domain-specific knowledge.
This challenge arises from the highly specialized nature of scientific instruments and methodologies, which are often underrepresented in LLMs' training data. \cite{bubeck2023sparks} demonstrated that while LLMs like GPT-4 excel in general academic tasks, they struggle with specialized scientific reasoning, particularly in physics and chemistry. 
Potential research directions include developing modular architectures for integrating domain-specific knowledge bases and tool interfaces, and fine-tuning LLMs on curated scientific datasets.
These approaches could enable LLMs to access domain-specific knowledge and interact effectively with specialized scientific tools, enhancing their capabilities in this setting.

\item \textit{Adaptive experimental design and hypothesis evolution}:
A significant challenge in scientific-focused agents is developing systems capable of long-term, iterative scientific investigations. Such agents must design experiments, interpret results, and refine hypotheses over extended periods while maintaining scientific rigor and avoiding biases. This challenge stems from the complex, multi-stage nature of scientific inquiry, which often involves repeated cycles of experimentation, analysis, and hypothesis adjustment.
Potential research directions to address this challenge include meta-learning frameworks enabling agents to improve experimental design and hypothesis refinement strategies across multiple investigations; and hierarchical planning algorithms for managing both short-term experimental steps and long-term scientific discovery objectives. 
% and ($iii$)~methods for ensuring scientific objectivity and mitigating confirmation bias in LLM-based iterative refinement processes.

\item \textit{Collaborative scientific reasoning}: 
Enabling collaborative scientific reasoning in AI agents is crucial for advancing scientific progress. Agents must build on their scientific knowledge, communicate hypotheses, engage in discourse, and critically judge peers' work. Current science agents struggle with deep critical analysis and identifying scientific flaws in AI-driven hypotheses and experimental designs \cite{birhane2023science}. Research opportunities include developing multi-agent systems simulating scientific communities, incorporating domain experts in the multi-agent refinement process, and creating benchmarks to enhance scientific discourse capabilities in science-focused agents.
\end{itemize}

%% OLD %%%%%%%%%%%%%%%%%%%%%%%
% Current work on scientific AI often treats models as passive tools rather than active agents pursuing discovery. There is a need to develop science-focused AI agents that can leverage broad scientific knowledge, engage in sustained reasoning, and autonomously pursue research goals.
% These agents should build on the impressive knowledge and reasoning capabilities of large language models, while incorporating additional structures for long-term planning, hypothesis tracking, and iterative experimentation. Key challenges include: ($i$)~Endowing agents with effective epistemic states to track hypotheses, evidence, and degrees of belief; ($i$)~Developing planning and meta-reasoning capabilities for long-term scientific investigations; ($ii$)~Creating interfaces between language models and external scientific tools or data sources; ($iii$)~Instilling appropriate inductive biases and priors from philosophy of science. 
% Inspiration can be drawn from cognitive architectures \cite{} and work on open-ended learning \cite{} toward discovery. The goal is to create AI systems that, like human scientists, can build on scientific prior knowledge from literature to identify promising research directions, reason about hypotheses, design and validate experiments, and iteratively build scientific understanding.
%%%%%%%%%%%%%%%%%%%%%%%%%%%%%%%%%%%%%%%%%

\subsection{Multi-modal Scientific Representations}
The landscape of scientific data is vast and diverse, encompassing far more than just textual information. While recent advancements in language models have significantly boosted our ability to process and reason with scientific literature, we must recognize that the majority of scientific data exists in forms quite different from natural language. From microscopy images to genomic sequences, from time series sensor data to structured databases and mathematical laws, scientific knowledge is inherently multi-modal \cite{topol2023artificial,wang2023scientific}.
This diversity presents both challenges and opportunities for AI-driven scientific discovery. The challenge lies in developing integrated representation learning techniques that can effectively capture and unify these varied scientific data types. The opportunity, however, is immense: by creating AI systems capable of reasoning across these diverse modalities, we can accelerate scientific discovery in unprecedented ways.

Representation learning offers the potential to distill complex, high-dimensional scientific data into more manageable continuous and low-dimensional forms. This is particularly crucial in scientific domains where high-quality data is limited or expensive to obtain through scientific experiments. By learning multi-modal robust representations with the help of pre-training techniques and synthetic simulation data, we can make more efficient use of limited data, potentially reducing the need for costly scientific experiments and accelerating the pace of discovery. Key directions in this line of research include: 

\begin{itemize}
\item \textit{Cross-modal scientific representation learning:}
Recent work has shown promising results in learning pre-trained joint representations across modalities for different scientific tasks. Notable successes include DrugCLIP \cite{gao2024drugclip} for joint representations of molecules and protein pockets in drug discovery, Text2Mol \cite{edwards2021text2mol} bridging natural language and molecular structures, ProtST \cite{xu2023protst} unifying protein sequences and biomedical text in proteomics, and SNIP \cite{meidani2024snip} linking mathematical expressions with numeric data. These advances demonstrate the potential of cross-modal learning to enhance scientific tasks by leveraging complementary information across modalities.
Despite these promising results, significant research opportunities remain ($i$)~\textit{Expanding} cross-modal representation learning to diverse and new scientific domains, ($ii$)~\textit{Enhancing} representation quality through recent integrated self-supervised and multi-modal pre-training; and ($iii$)~Developing unified, modality-agnostic frameworks adaptable to heterogeneous scientific data types.

\item \textit{Latent space scientific hypothesis search}:
% \textit{Representation-Based Scientific Hypothesis Search}
Many scientific discovery tasks involve searching through vast, combinatorial spaces of candidates. Current approaches to these problems often rely on evolutionary search or heuristic methods, which can be computationally expensive and inefficient \cite{sadybekov2023computational,schmidt2009symbolic}.
Recent advances in representation learning offer a promising alternative: conducting scientific hypothesis optimization in learned latent spaces.
By moving the search process into the latent space, we can potentially make the exploration of the hypothesis space more efficient and effective.
This approach has shown potential across various domains, from drug discovery \cite{gao2024drugclip} to equation discovery \cite{meidani2024snip}, molecular design \cite{abeer2024multi,zheng2023desirable}, and protein engineering \cite{castro2022transformer,jumper2021highly}.
This emerging research direction has significant potential for scientific discovery. Future research avenues include ($i$)~Integrating domain expert knowledge or feedback into the representations and discovery process, ($ii$)~Enhancing interpretability of representations for scientific validation, and ($iii$)~Advancing optimization techniques for nontrivial discovery objectives and more flexible hypothesis search in the latent space.

%%%%%%% LITERATURE %%%%%%%%%%%%%
% Recent work has demonstrated the potential of this approach across various scientific domains. In drug discovery, DrugCLIP \cite{gao2024drugclip} achieved state-of-the-art performance by searching in a joint latent space of molecules and protein pockets. For equation discovery, SNIP \cite{meidani2024snip} has effectively shown to discover scientific laws by exploring a latent space bridging mathematical expressions and numeric data. In molecular design, researchers have developed methods for multi-objective optimization in latent spaces \cite{abeer2024multi,zheng2023desirable}, while in protein engineering, models like ReLSO \cite{castro2022transformer} have shown how latent space search can accelerate the design of novel proteins.

% There are still great potentials for research in this direction towards scientific discovery. As scientific discovery usually build on the knowledge of domain experts and in scientific interpretability and understandability by domains experts is needed for scientific validation, there's potential for more research on these generative latent space interpretability or the inclusion of domain expert feedback int other search process with the help of gradient-free optimization for scientific hypothesis search in the richer space of representations.  
%%%%%%%%%%%%%%%%%%%%%%%%%%%%%%

\item \textit{Multi-modal scientific  reasoning frameworks:}
The advancement of AI-driven scientific discovery hinges on developing systems capable of multi-modal scientific reasoning. 
% Scientific knowledge inherently spans diverse modalities, including text, images, graphs, tabular, and structured data. Effective reasoning often requires seamless integration of information across these modalities. 
Recent works have shown promising results in this direction. 
For example, multi-modal retrieval augmented generation (RAG) systems have demonstrated potential in leveraging LLMs for scientific discovery \cite{park2024leveraging}. Models like GIT-Mol \cite{liu2024git} showcase the integration of visual, textual, and graph reasoning for molecular discovery. In materials science, approaches combining textual reasoning with structural data have also shown promise in predicting material properties and guiding synthesis \cite{miret2024llms}.
% multi-modal retrieval augmented generation (RAG) systems have recently shown great potential in leveraging LLMs for scientific discovery \cite{park2024leveraging}. Models like GIT-Mol \cite{liu2024git} demonstrate the potential of combining visual, textual, and graph reasoning for molecular discovery. In materials science, approaches that integrate textual reasoning with structural data have shown promise in predicting material properties and guiding synthesis \cite{miret2024llms}.
However, comprehensive multi-modal scientific reasoning frameworks remain an open challenge. Such frameworks must effectively integrate reasoning across diverse data types. While studies like \cite{scienceqa} have shown improved scientific question-answering through combined text and image contexts, further research is needed to explore the impact of other modalities such as numerical or tabular data, and symbolic mathematical theories on scientific discovery tasks.
% However, comprehensive multi-modal scientific reasoning frameworks remain an open challenge. Such frameworks would need to effectively integrate reasoning across diverse data types. For example, \cite{scienceqa} demonstrated that combining scientific text and image contexts boosts reasoning of scientific question-answering tasks. There's still potential to further study the impact of other modalities such as tabular data, numerical data, as well as symbolic mathematics in the reasoning required for scientific discovery tasks. 

%%%%%%%%%%%%%%
% Multi-modal retrieval augmented generation (RAG) systems have recently shown great potential in leveraging LLMs for scientific discovery. 
% For instance, \cite{bran2023chemcrow} demonstrated how a tool-augmented GPT-4 system could perform end-to-end experimental planning and execution for chemical synthesis based on user queries. This approach combines the reasoning capabilities of LLMs with domain-specific tools and databases, enabling more robust scientific reasoning.
% \cite{miret2024llms} found that even advanced LLMs like GPT-4 struggle with complex numerical problems and 3D structural reasoning in materials science. Ensuring that multi-modal reasoning frameworks are properly grounded in fundamental scientific principles is an open research direction.
%%%%%%%%%%%%%%

\item \textit{Transfer learning in scientific domains}:
Transfer learning offers great potential to accelerate scientific discovery, particularly in domains where data is limited or expensive to obtain. Recent studies have demonstrated its efficacy across various scientific fields: In drug discovery, models pre-trained on large synthetic chemical databases have shown improved performance in predicting properties of novel compounds \cite{gao2024drugclip}. In materials science, transfer learning from simulated data to real-world experiments has also accelerated the discovery of new materials with desired properties \cite{chen2024knowledge}.
However, the application of transfer learning in scientific domains presents unique challenges due to the high specificity of scientific knowledge and potential domain shift between source and target tasks. Advancing these capabilities could unlock new avenues for cross-disciplinary discoveries and accelerate progress in data-scarce scientific domains.

\end{itemize}

\subsection{Theory and Data Unification}
Scientific discovery typically involves a complex interplay between theoretical reasoning, empirical observation, and mathematical modeling. 
However, most existing AI approaches to scientific tasks focus on just one of these aspects. There is a pressing need for unified frameworks that integrate logical and mathematical reasoning, formal theorem proving, data-driven modeling, experimental design, and causal inference. This integration is challenging but critical for capturing the full scientific discovery process.
Recent advances in LLMs have shown promising results in both theorem-proving and data-driven scientific modeling.
For instance, LLMs have demonstrated promising capabilities in automated theorem-proving and formal mathematical derivations from natural language problems \cite{yang2024leandojo,jiang2023draft}. On the data-driven side, \cite{shojaee2024llm,llmsim-bilevel} have shown success in discovering equation hypotheses from data with the help of LLM-based program search. However, these approaches largely operate in isolation, and there is a significant gap in unifying these capabilities to mirror the holistic nature of scientific inquiry.  Key challenges and research directions include:

\begin{itemize}
\item \textit{Generating derivable hypotheses from empirical observations}: Developing methods that can not only discover patterns in data but also produce rigorous mathematical derivations of these findings is crucial for ensuring the reliability and generalizability of AI-driven scientific discoveries to out-of-distribution data. Derivable theoretical results provide a level of confidence and understanding that goes beyond mere empirical correlation. Recent work, such as the AI-Descartes system \cite{cornelio2023combining}, has shown promise by combining equation discovery tools (known as symbolic regression) with automated logical reasoning. 
However, integrating logical reasoning and data-driven frameworks that are adaptable across scientific discovery tasks still remains an open challenge. Research opportunities exist to automate proof verification, incorporate expert feedback, and embed derivability constraints in data-driven discovery algorithms. 

\item \textit{Combining symbolic and neural approaches}: How can we effectively integrate the strengths of symbolic reasoning (e.g., logical deduction, formal proofs) with the flexibility and learning capabilities of neural networks? Recent work on neuro-symbolic AI \cite{garcez2023neurosymbolic,sheth2023neurosymbolic} provides promising directions, but challenges remain in scaling these approaches to more complex settings and scientific tasks. Developing hybrid architectures that can transition between symbolic and neural representations is helpful in capturing the full spectrum of scientific reasoning.

% \item \textit{Leveraging theoretical knowledge as constraints}: One of the key challenges is how to use existing scientific theories and laws to guide and constrain data-driven discovery processes. This could involve incorporating domain knowledge into neural network architecture designs or using theoretical constraints in optimization objectives. For example, physics-informed \cite{raissi2019physics} and lagrangian neural networks \cite{cranmer2020lagrangian}  exemplify this approach, integrating physical symmetries and conservation laws into neural models. Expanding on this area of research to cover a broader range of scientific domains and theoretical principles is still an important research direction.

\item \textit{Reasoning discovery uncertainty in formal frameworks}: Scientific discoveries often involve uncertainties and probabilities, yet formal logical frameworks struggle to incorporate these aspects. Developing frameworks that can handle probabilistic reasoning while maintaining rigorous deduction capabilities is crucial for advancing AI-driven scientific discovery. Recent work, such as probabilistic logic systems \cite{de2015probabilistic,problog}, and neuro-symbolic programming \cite{ahmed2022semantic} has made progress in this direction. However, significant challenges remain for the use of these approaches in scientific discovery, including scalability to large-scale scientific problems, and expressiveness to capture complex scientific theories in specific scientific domains.

% This is a very important problem in the direction of scientific discovery. Having a more comprehensive framework integrating logical derivation and data-driven modeling that can be adapted to various scientific discovery tasks is still an open research problem. There are still research opportunities for automating proof verification for intermediate discoveries, incorporating human scientist theoretical feedback in the process, and embedding derivability constraints in discovery algorithms.
%  Advances in this line of research could yield AI systems capable of contributing to rigorous scientific theory development.

% My view 
% However, there is still need for more comprehensive frameworks integrating logical derivation and data-driven modeling and further exploration of this line of research in other scientific discovery tasks. potential future research avenues include  integrating domain theorist feedback in the data-driven discoveries, focusing on the decomposition of models or tasks into verifiable components, developing automated proof-verfication systems for intermediate data-driven discovery process, and integrating derivability constraints into discovery algorithms. Advances in this line of research could yield AI systems capable of contributing to rigorous scientific theory development.
\end{itemize}

% , leading to breakthroughs in crucial areas like clean energy, medicine, and fundamental physics.

\section{Conclusion}
Developing unified AI systems for scientific discovery is an ambitious goal, but one with substantial potential impact. Success could dramatically accelerate progress across diverse scientific disciplines. This paper has outlined current progress as well as several key research challenges and opportunities toward this vision, including developing science-focused AI agents, creating improved benchmarks, advancing multimodal representations, and unifying diverse modes of scientific reasoning. Tackling these challenges will require collaboration between AI researchers, scientists across domains, and philosophers of science.
While fully autonomous AI scientists may still be far off, nearer-term progress could produce powerful AI assistants to augment human scientific capabilities. Such tools could help scientists navigate the ever-growing scientific literature, brainstorm ideas, generate novel hypotheses, design experiments, and find unexpected patterns in complex experimental data.
By pursuing this research agenda, the machine learning and AI community has an opportunity to develop systems that do not just automate product-related tasks, but actively push forward the frontiers of human scientific knowledge. The path will be challenging, but the potential rewards - both scientific and technological - are immense.

\appendix

\bibliography{aaai25}

\end{document}